\documentclass[11pt,a4paper]{article}
\usepackage[hyperref]{emnlp2018}
\usepackage{times}
\usepackage{latexsym}
\usepackage{graphicx}
\usepackage{url}
\usepackage{tabularx}
\usepackage{amsmath,amssymb,amstext}

\usepackage{url}

\def\*#1{\mathbf{#1}}
\DeclareMathOperator*{\argmax}{arg\,max}
\aclfinalcopy 

\newcommand{\B}{\textsc{BanditSum }}
\newcommand{\Bnospace}{\textsc{BanditSum}}

\title{\textsc{BanditSum}: Extractive Summarization as a Contextual Bandit}
\author{Yue Dong\thanks{\;\;Equal contribution. 
  } \\
  Mila/McGill University \\ 
  {\tt yue.dong2}\\{\tt@mail.mcgill.ca} \\
  \And
  Yikang Shen$^{*}$ \\
  Mila/University of Montr\'eal \\ 
  {\tt yi-kang.shen}\\{\tt@umontreal.ca} \\
  \AND
  Eric Crawford \\
  Mila/McGill University\\
  {\tt eric.crawford}\\{\tt@mail.mcgill.ca} \\
  \And
  Herke van Hoof \\
  University of Amsterdam\\
  {\tt h.c.vanhoof}\\{\tt@uva.nl} \\
  \And
  Jackie C.K. Cheung \\
  Mila/McGill University\\
  {\tt jcheung}\\{\tt@cs.mcgill.ca}
  }

\date{}

\begin{document}
\maketitle
\begin{abstract}
In this work, we propose a novel method for training neural networks to perform single-document extractive summarization without heuristically-generated extractive labels. We call our approach \B as it treats extractive summarization as a contextual bandit (CB) problem, where the model receives a document to summarize (the context), and chooses a sequence of sentences to include in the summary (the action). A policy gradient reinforcement learning algorithm is used to train the model to select sequences of sentences that maximize ROUGE score. We perform a series of experiments demonstrating that \B is able to achieve ROUGE scores that are better than or comparable to the state-of-the-art for extractive summarization, and converges using significantly fewer update steps than competing approaches. In addition, we show empirically that \B performs significantly better than competing approaches when good summary sentences appear late in the source document\footnote{Our code can be found at \url{https://github.com/yuedongP/BanditSum}}.

\end{abstract}

\section{Introduction}
Single-document summarization methods can be divided into two categories: extractive and abstractive. Extractive summarization systems form summaries by selecting and copying text snippets from the document, while abstractive methods aim to generate concise summaries with paraphrasing. This work is primarily concerned with extractive summarization. Though abstractive summarization methods have made strides in recent years, extractive techniques are still very attractive as they are simpler, faster, and more reliably yield semantically and grammatically correct sentences.

Many extractive summarizers work by selecting sentences from the input document \citep{tra_ext1_luhn1958automatic,tra_ext3_mihalcea2004textrank,tra_ext5_wong2008extractive,ext1_kaageback2014extractive,ext2_2015Yin,ext3_cao2015learning,yasunaga2017graph}. Furthermore, a growing trend is to frame this sentence selection process as a sequential binary labeling problem, where binary inclusion/exclusion labels are chosen for sentences one at a time, starting from the beginning of the document, and decisions about later sentences may be conditioned on decisions about earlier sentences. Recurrent neural networks may be trained with stochastic gradient ascent to maximize the likelihood of a set of ground-truth binary label sequences \citep{ext4_cheng2016neural,ext5_summarunner}.
However, this approach has two well-recognized disadvantages. First, it suffers from exposure bias, a form of mismatch between training and testing data distributions which can hurt performance \citep{Ranzato2015SequenceLT, rl1_bahdanau+al-2017-actorcritic-iclr, abs5_paulus2017deep}. Second, extractive labels must be generated by a heuristic, as summarization datasets do not generally include ground-truth extractive labels; the ultimate performance of models trained on such labels is thus fundamentally limited by the quality of the heuristic.


An alternative to maximum likelihood training is to use reinforcement learning to train the model to directly maximize a measure of summary quality, such as the ROUGE score between the generated summary and a ground-truth abstractive summary \citep{DBLP:conf/aaai/WuH18}. This approach has become popular because it avoids exposure bias, and directly optimizes a measure of summary quality. However, it also has a number of downsides. For one, the search space is quite large: for a document of length $T$, there are $2^T$ possible extractive summaries. This makes the exploration problem faced by the reinforcement learning algorithm during training very difficult. Another issue is that due to the sequential nature of selection, the model is inherently biased in favor of selecting earlier sentences over later ones, a phenomenon which we demonstrate empirically in Section \ref{sec:discussion}. The first issue can be resolved to a degree using either a cumbersome maximum likelihood-based pre-training step (using heuristically-generated labels) \citep{DBLP:conf/aaai/WuH18}, or placing a hard upper limit on the number of sentences selected. The second issue is more problematic, as it is inherent to the sequential binary labeling setting.


In the current work, we introduce \textsc{BanditSum}, a novel method for training neural network-based extractive summarizers with reinforcement learning. This method does away with the sequential binary labeling setting, instead formulating extractive summarization as a contextual bandit. This move greatly reduces the size of the space that must be explored, removes the need to perform supervised pre-training, and prevents systematically privileging earlier sentences over later ones. Although the strong performance of Lead-3 indicates that good sentences often occur early in the source article, we show in Sections \ref{sec:results} and \ref{sec:discussion} that the contextual bandit setting greatly improves model performance when good sentences occur late without sacrificing performance when good sentences occur early.

Under this reformulation, \B takes the document as input and outputs an \textit{affinity} for each of the sentences therein. An affinity is a real number in $[0, 1]$ which quantifies the model's propensity for including a sentence in the summary. These affinities are then used in a process of repeated sampling-without-replacement which does not privilege earlier sentences over later ones. \B is free to process the document as a whole before yielding affinities, which permits affinities for different sentences in the document to depend on one another in arbitrary ways. In our technical section, we show how to apply policy gradient reinforcement learning methods to this setting. 

The contributions of our work are as follows:
\begin{itemize}
\item We propose a theoretically grounded method, based on the contextual bandit formalism, for training neural network-based extractive summarizers with reinforcement learning. Based on this training method, we propose the \B system for extractive summarization.

\item We perform experiments demonstrating that \B obtains state-of-the-art performance on a number of datasets and requires significantly fewer update steps than competing approaches.

\item We perform human evaluations showing that in the eyes of human judges, summaries created by \B are less redundant and of higher overall quality than summaries created by competing approaches.

\item We provide evidence, in the form of experiments in which models are trained on subsets of the data, that the improved performance of \B over competitors stems in part from better handling of summary-worthy sentences that come near the end of the document (see Section~\ref{sec:discussion}).
\end{itemize}

\section{Related Work\label{sec:related-work}}
Extractive summarization has been widely studied in the past. Recently, neural network-based methods have been gaining popularity over classical methods \citep{tra_ext1_luhn1958automatic,tra_ext2_gong2001generic, tra_ext4_conroy2001text,tra_ext3_mihalcea2004textrank,tra_ext5_wong2008extractive}, as they have demonstrated stronger performance on large corpora. Central to the neural network-based models is the encoder-decoder structure. These models typically use either a convolution neural network \cite{cnn1_kalchbrenner2014convolutional,cnn2_kim2014convolutional,ext2_2015Yin,ext3_cao2015learning}, a recurrent neural network \cite{rnn2_chung2014gru,ext4_cheng2016neural,ext5_summarunner}, or a combination of the two \cite{DBLP:Narayan/2018,DBLP:conf/aaai/WuH18} to create sentence and document representations, using word embeddings \cite{we1_mikolov2013efficient,we2_pennington2014glove} to represent words at the input level. These vectors are then fed into a decoder network to generate the output summary. 

The use of reinforcement learning (RL) in extractive summarization was first explored by \citet{rl_ryang2012frameworkRL}, who proposed to use the TD($\lambda$) algorithm to learn a value function for sentence selection. \citet{rl_rioux2014fear} improved this framework by replacing the learning agent with another TD($\lambda$) algorithm. However, the performance of their methods was limited by the use of shallow function approximators, which required performing a fresh round of reinforcement learning for every new document to be summarized. The more recent work of \citet{abs5_paulus2017deep} and \citet{DBLP:conf/aaai/WuH18} use reinforcement learning in a sequential labeling setting to train abstractive and extractive summarizers, respectively, while \citet{chen2018abstractive} combines both approaches, applying abstractive summarization to a set of sentences extracted by a pointer network \citep{vinyals2015pointer} trained via REINFORCE.
However, pre-training with a maximum likelihood objective is required in all of these models.

The two works most similar to ours are \citet{yao2018deep} and \citet{DBLP:Narayan/2018}. \citet{yao2018deep} recently proposed an extractive summarization approach based on deep Q learning, a type of reinforcement learning. However, their approach is extremely computationally intensive (a minimum of 10 days before convergence), and was unable to achieve ROUGE scores better than the best maximum likelihood-based approach. \citet{DBLP:Narayan/2018} uses a cascade of filters in order to arrive at a set of candidate extractive summaries, which we can regard as an approximation of the true action space. They then use an approximation of a policy gradient method to train their neural network to select summaries from this approximated action space. In contrast, \B samples directly from the true action space, and uses exact policy gradient parameter updates.

\section{Extractive Summarization as a Contextual Bandit \label{sec:technical}}
Our approach formulates extractive summarization as a contextual bandit which we then train an agent to solve using policy gradient reinforcement learning. A bandit is a decision-making formalization in which an agent repeatedly chooses one of several actions, and receives a reward based on this choice. The agent's goal is to quickly learn which action yields the most favorable distribution over rewards, and choose that action as often as possible. In a \textit{contextual} bandit, at each trial, a context is sampled and shown to the agent, after which the agent selects an action and receives a reward; importantly, the rewards yielded by the actions may depend on the sampled context. The agent must quickly learn which actions are favorable in which contexts. Contextual bandits are a subset of Markov Decision Processes in which every episode has length one.

Extractive summarization may be regarded as a contextual bandit as follows. Each document is a context, and each ordered subset of a document's sentences is a different action. Formally, assume that each context is a document $d$ consisting of sentences $s = (s_1, \dots, s_{N_d})$, and that each action is a length-$M$ sequence of unique sentence indices $i = (i_1, \dots, i_M)$ where $i_t \in \{1, \dots, N_d\}$, $i_t \neq i_{t'}$ for $t \neq t'$, and $M$ is an integer hyper-parameter. For each $i$, the extractive summary induced by $i$ is given by $(s_{i_1}, \dots, s_{i_M})$. An action $i$ taken in context $d$ is given a reward $R(i, a)$, where $a$ is the gold-standard abstractive summary that is paired with document $d$, and $R$ is a scalar reward function quantifying the degree of match between $a$ and the summary induced by $i$.

A policy for extractive summarization is a neural network $p_\theta(\cdot | d)$, parameterized by a vector $\theta$, which, for each input document $d$, yields a probability distribution over index sequences. Our goal is to find parameters $\theta$ which cause $p_\theta(\cdot|d)$ to assign high probability to index sequences that induce extractive summaries that a human reader would judge to be of high-quality. We achieve this by maximizing the following objective function with respect to parameters $\theta$:
\begin{align}
    J(\theta) = E\left[R(i, a)\right]\label{eqn:objective}
\end{align}
where the expectation is taken over documents $d$ paired with gold-standard abstractive summaries $a$, as well as over index sequences $i$ generated according to $p_\theta(\cdot | d)$.

\subsection{Policy Gradient Reinforcement Learning}
Ideally, we would like to maximize \eqref{eqn:objective} using gradient ascent. However, the required gradient cannot be obtained using usual techniques (e.g. simple backpropagation) because $i$ must be discretely sampled in order to compute $R(i, a)$.

Fortunately, we can use the likelihood ratio gradient estimator from reinforcement learning and stochastic optimization \cite{williams1992simple, sutton2000policy}, which tells us that the gradient of this function can be computed as:
\begin{align}
    \nabla_\theta J(\theta) = E\left[ \nabla_\theta \log p_\theta(i | d) R(i, a) \right] \label{eqn:gradient}
\end{align}
where the expectation is taken over the same variables as \eqref{eqn:objective}.

Since we typically do not know the exact document distribution and thus cannot evaluate the expected value in \eqref{eqn:gradient}, we instead estimate it by sampling. We found that we obtained the best performance when, for each update, we first sample one document/summary pair $(d, a)$, then sample $B$ index sequences $i^1, \dots, i^B$ from $p_\theta(\cdot | d)$, and finally take the empirical average:
\begin{align}
    \nabla_\theta J(\theta) \approx \frac{1}{B} \sum_{b=1}^{B}\nabla_\theta \log p_\theta(i^b | d) R(i^b, a) \label{eqn:estimated-gradient}
\end{align}
This overall learning algorithm can be regarded as an instance of the REINFORCE policy gradient algorithm \citep{williams1992simple}.

\subsection{Structure of $p_\theta(\cdot|d)$}
There are many possible choices for the structure of $p_\theta(\cdot | d)$; we opt for one that avoids privileging early sentences over later ones. We first decompose $p_\theta(\cdot | d)$ into two parts: $\pi_\theta$, a deterministic function which contains all the network's parameters, and $\mu$, a probability distribution parameterized by the output of $\pi_\theta$. Concretely:
\begin{align}
    p_\theta(\cdot | d) = \mu(\cdot | \pi_\theta(d))
\end{align}
Given an input document $d$, $\pi_\theta$ outputs a real-valued vector of \textit{sentence affinities} whose length is equal to the number of sentences in the document (i.e. $\pi_\theta(d) \in \mathbb{R}^{N_d}$) and whose elements fall in the range $[0, 1]$. The $t$-th entry $\pi(d)_t$ may be roughly interpreted as the network's propensity to include sentence $s_t$ in the summary of $d$.

Given sentence affinities $\pi_\theta(d)$, $\mu$ implements a process of repeated sampling-without-replacement. This proceeds by repeatedly normalizing the set of affinities corresponding to sentences that have not yet been selected, thereby obtaining a probability distribution over unselected sentences, and sampling from that distribution to obtain a new sentence to include. This normalize-and-sample step is repeated $M$ times, yielding $M$ unique sentences to include in the summary.

At each step of sampling-without-replacement, we also include a small probability $\epsilon$ of sampling uniformly from all remaining sentences. This is used to achieve adequate exploration during training, and is similar to the $\epsilon$-greedy technique from reinforcement learning.

Under this sampling scheme, we have the following expression for $p_\theta(i|d)$:
\begin{align}
\prod_{j=1}^M \left( \frac{\epsilon}{N_d - j + 1} + \frac{(1-\epsilon)\pi(d)_{i_j}}{z(d) - \sum_{k=1}^{j-1} \pi(d)_{i_k}} \right)
\end{align}
where $z(d) = \sum_t\pi(d)_t$. For index sequences that have length different from $M$, or that contain duplicate indices, we have $p_\theta(i|d) = 0$. Using this expression, it is straightforward to use automatic differentiation software to compute $\nabla_\theta \log p_\theta(i|d)$, which is required for the gradient estimate in \eqref{eqn:estimated-gradient}.

\subsection{Baseline for Variance Reduction} \label{subsec:baseline}
Our sample-based gradient estimate can have high variance, which can slow the learning. One potential cause of this high variance can be seen by inspecting \eqref{eqn:estimated-gradient}, and noting that it basically acts to change the probability of a sampled index sequence to an extent determined by the reward $R(i, a)$. However, since ROUGE scores are always positive, the probability of every sampled index sequence is increased, whereas intuitively, we would prefer to decrease the probability of sequences that receive a comparatively low reward, even if it is positive. This can be remedied by the introduction of a so-called baseline which is subtracted from all rewards.

Using a baseline $\overline{r}$, our sample-based estimate of $\nabla_\theta J(\theta)$ becomes:
\begin{align}
    \frac{1}{B} \sum_{i=1}^{B}\nabla_\theta \log p_\theta(i^b | d) (R(i^b, a) - \overline{r})\label{eqn:estimated-gradient-baseline}
\end{align}
It can be shown that the introduction of $\overline{r}$ does not bias the gradient estimator and can significantly reduce its variance if chosen appropriately \citep{sutton2000policy}. 

There are several possibilities for the baseline, including the long-term average reward and the average reward across different samples for one document-summary pair. We choose an approach known as self-critical reinforcement learning, in which the test-time performance of the current model is used as the baseline \citep{Ranzato2015SequenceLT,rennie2017self,abs5_paulus2017deep}. More concretely, after sampling the document-summary pair $(d, a)$, we greedily generate an index sequence using the current parameters $\theta$:
\begin{align}
    i_\textit{greedy} = \argmax_i p_\theta(i | d)
\end{align}
and calculate the baseline for the current update as $\overline{r} = R(i_\textit{greedy}, a)$. This baseline has the intuitively satisfying property of only increasing the probability of a sampled label sequence when the summary it induces is better than what would be obtained by greedy decoding. 

\subsection{Reward Function}
A final consideration is a concrete choice for the reward function $R(i, a)$. Throughout this work we use:
\begin{multline}
R(i, a) = \frac{1}{3} (\text{ROUGE-1}_f(i, a) + {} \\
                     \text{ROUGE-2}_f(i, a) + 
                     \text{ROUGE-L}_f(i, a)).
\end{multline}
The above reward function optimizes the average of all the ROUGE variants \cite{eva1_lin:2004:ACLsummarization} while balancing precision and recall.

\section{Model\label{sec:model}}
In this section, we discuss the concrete instantiations of the neural network $\pi_\theta$ that we use in our experiments. We break $\pi_\theta$ up into two components: a document encoder $f_{\theta1}$, which outputs a sequence of sentence feature vectors $(h_1, \dots, h_{N_d})$ and a decoder $g_{\theta2}$ which yields sentence affinities:
\begin{align}
    h_1, \dots, h_{N_d} &= f_{\theta1}(d)\\
    \pi_\theta(d) &= g_{\theta2}(h_1, \dots, h_{N_d})
\end{align}

\noindent \textbf{Encoder.} Features for each sentence in isolation are first obtained by applying a word-level Bidirectional Recurrent Neural Network (BiRNN) to the embeddings for the words in the sentence, and averaging the hidden states over words. A separate sentence-level BiRNN is then used to obtain a representations $h_i$ for each sentence in the context of the document.


\noindent \textbf{Decoder.} A multi-layer perceptron is used to map from the representation $h_t$ of each sentence through a final sigmoid unit to yield sentence affinities $\pi_\theta(d)$.

The use of a bidirectional recurrent network in the encoder is crucial, as it allows the network to process the document as a whole, yielding representations for each sentence that take all other sentences into account. This procedure is necessary to deal with some aspects of summary quality such as redundancy (avoiding the inclusion of multiple sentences with similar meaning), which requires the affinities for different sentences to depend on one another. For example, to avoid redundancy, if the affinity for some sentence is high, then sentences which express similar meaning should have low affinities.

\section{Experiments\label{sec:experiments}}
In this section, we discuss the setup of our experiments. We first discuss the corpora that we used and our evaluation methodology. We then discuss the baseline methods against which we compared, and conclude with a detailed overview of the settings of  the model parameters. 
\subsection{Corpora}
Three datasets are used for our experiments: the CNN, the Daily Mail, and combined CNN/Daily Mail \citep{data1_hermann2015teaching,data2_nallapati2016abstractive}. We use the standard split of \citet{data1_hermann2015teaching} for training, validating, and testing and the same setting without \textit{anonymization} on the three corpus as \citet{abs4_SeeLM17}. The Daily Mail corpus has 196,557 training documents, 12,147 validation documents and 10,397 test documents; while the CNN corpus has 90,266/1,220/1,093 documents, respectively. 



\subsection{Evaluation}
The models are evaluated based on ROUGE \cite{eva1_lin:2004:ACLsummarization}.
We obtain our ROUGE scores using the standard pyrouge
package\footnote{\url{https://pypi.python.org/pypi/pyrouge/0.1.3}} for the test set evaluation and a faster python implementation of the ROUGE metric\footnote{We use the modified version based on \url{https://github.com/pltrdy/rouge}}
for training and evaluating on the validation set. We report the F1 scores of ROUGE-1, ROUGE-2, and ROUGE-L, which compute the uniform, bigram, and longest common subsequence overlapping with the reference summaries.

\subsection{Baselines}
We compare \B with other extractive methods including: the Lead-3 model, SummaRuNNer \citep{ext5_summarunner}, Refresh \citep{DBLP:Narayan/2018}, RNES \citep{DBLP:conf/aaai/WuH18}, DQN \citep{yao2018deep}, and NN-SE \citep{ext4_cheng2016neural}.  The Lead-3 model simply produces the leading three sentences of the document as the summary. 

\subsection{Model Settings}
We use 100-dimensional Glove embeddings \citep{we2_pennington2014glove} as our embedding initialization. We do not limit the sentence length, nor the maximum number of sentences per document. We use one-layer BiLSTM for word-level RNN, and two-layers BiLSTM for sentence-level RNN. The hidden state dimension is 200 for each direction on all LSTMs. For the decoder, we use a feed-forward network with one hidden layer of  dimension 100. 

During training, we use Adam \citep{adam_kingma2014adam} as the optimizer with the learning rate of $5e^{-5}$, beta parameters $(0, 0.999)$, and a weight decay of $1e^{-6}$, to maximize the objective function defined in equation \eqref{eqn:objective}. We employ gradient clipping of 1 to regularize our model. At each iteration, we sample $B = 20$ times to estimate the gradient defined in equation \ref{eqn:estimated-gradient}. For our system, the reported performance is obtained within two epochs of training.


At the test time, we pick sentences sorted by the predicted probabilities until the length limit is reached. The full-length ROUGE F1 score is used as the evaluation metric. For $M$, the number of sentences selected per summary, we use a value of 3, based on our validation results as well as on the settings described in \citet{ext5_summarunner}.

\section{Experiment Results\label{sec:results}}
In this section, we present  quantitative results from the ROUGE evaluation and qualitative results based on human evaluation. In addition, we demonstrate the stability of our RL model by comparing the validation curve of \B with SummaRuNNer \citep{ext5_summarunner} trained with a maximum likelihood objective. 
\subsection{Rouge Evaluation}


\begin{table}[!h]
\footnotesize
\centering
\begin{tabular}{|l|l|l|l|}
\hline
Model & \multicolumn{3}{l|}{ROUGE} \\ 
\hline
 & 1 & 2 & L \\ 
 \hline
Lead\citep{DBLP:Narayan/2018} & 39.6 & 17.7 & 36.2 \\ 
Lead-3(ours) & 40.0 & 17.5 & 36.2 \\
SummaRuNNer & 39.6 & 16.2 & 35.3  \\
DQN & 39.4	& 16.1 &	35.6 \\ 
Refresh & 40.0 &18.2 &36.6\\
RNES w/o coherence & 41.3 & \textbf{18.9} & \textbf{37.6}\\ 
\hline
\B  & \textbf{41.5} & 18.7 & \textbf{37.6} \\ 
\hline
\end{tabular}
\caption{Performance comparison of different extractive summarization models on the combined CNN/Daily Mail test set using full-length F1. }
\label{table:results_cnn}
\end{table}

\begin{table}[!h]
\small
\centering
\setlength\tabcolsep{4 pt}
\begin{tabular}{|l|l|l|l|l|l|l|}
\hline
Model & \multicolumn{3}{l|}{CNN} & \multicolumn{3}{l|}{Daily Mail} \\ \hline
 & 1 & 2 & L & 1 & 2 & L   \\ \hline
Lead-3 & 28.8 & 11.0 & 25.5 & 41.2 & 18.2 & 37.3 \\
NN-SE & 28.4 & 10.0 & 25.0 & 36.2 & 15.2 & 32.9 \\
Refresh & 30.4 & \textbf{11.7} & 26.9& 41.0 & 18.8 & 37.7 \\ 
\B & \textbf{30.7} & 11.6 & \textbf{27.4} & \textbf{42.1} & \textbf{18.9} &  \textbf{38.3}\\  
\hline
\end{tabular}
\caption{The full-length ROUGE F1 scores of various extractive models on the CNN and the Daily Mail test set separately.}
\label{table:daily_mail}
\end{table}

We present the results of comparing \B to several baseline algorithms\footnote{
Due to different pre-processing methods and different numbers of selected sentences, several papers report different Lead scores \citep{DBLP:Narayan/2018,abs4_SeeLM17}. 
We use the test set provided by \citet{DBLP:Narayan/2018}. Since their Lead score is a combination of Lead-3 for CNN and Lead-4 for Daily Mail, we recompute the Lead-3 scores for both CNN and Daily Mail with the preprocessing steps used in \citet{abs4_SeeLM17}. Additionally, our results are not directly comparable to results based on the anonymized dataset used by \citet{ext5_summarunner}.} 
on the CNN/Daily Mail corpus in Tables~\ref{table:results_cnn} and \ref{table:daily_mail}. 
Compared to other extractive summarization systems, \B achieves performance that is significantly better than two RL-based approaches, Refresh \citep{DBLP:Narayan/2018} and DQN \citep{yao2018deep}, as well as SummaRuNNer, the state-of-the-art maximum liklihood-based extractive summarizer \cite{ext5_summarunner}. \B performs a little better than RNES \citep{DBLP:conf/aaai/WuH18} in terms of ROUGE-1 and slightly worse in terms of ROUGE-2. However, RNES requires pre-training with the maximum likelihood objective on heuristically-generated extractive labels; in contrast, \B is very light-weight and converges significantly faster. We discuss the advantage of framing the extractive summarization based on the contextual bandit (\textsc{BanditSum}) over the sequential binary labeling setting (RNES) in the discussion Section~\ref{sec:discussion}. 



We also noticed that different choices for the policy gradient baseline (see Section~\ref{subsec:baseline}) in \B affect learning speed, but do not significantly affect asymptotic performance. Models trained with an average reward baseline learned most quickly, while models trained with three different baselines (greedy, average reward in a batch, average global reward) all perform roughly the same after training for one epoch. Models trained without a baseline were found to under-perform other baseline choices by about 2 points of ROUGE score on average.  

\subsection{Human Evaluation}
We also conduct a qualitative evaluation to understand the effects of the improvements introduced in \B on human judgments of the generated summaries. To assess the effect of training with RL rather than maximum likelihood, in the first set of human evaluations we compare \B with the state-of-the-art maximum likelihood-based model SummaRuNNer. To evaluate the importance of using an exact, rather than approximate, policy gradient to optimize ROUGE scores, we perform another human evaluation comparing \B and Refresh, an RL-based method that uses the an approximation of the policy gradient.

We follow a human evaluation protocol similar to the one used in \citet{DBLP:conf/aaai/WuH18}. Given a set of $N$ documents, we ask $K$ volunteers to evaluate the summaries extracted by both systems. For each document, a reference summary, and a pair of randomly ordered extractive summaries (one generated by each of the two models) is presented to the volunteers. They are asked to compare and rank the extracted summaries along three dimensions: overall, coverage, and non-redundancy.

\begin{table}[h!]
\small
\begin{center}
\begin{tabularx}{\columnwidth}{|l|X|X|X|}
\hline
Model & Overall & Coverage & Non-Redundancy \\ \hline
SummaRuNNer & 1.67 &\textbf{1.46}  &1.70  \\ \hline
\B & \textbf{1.33} & 1.54 & \textbf{1.30} \\ \hline
\end{tabularx}
\end{center}
\caption{Average rank of human evaluation based on 5 participants who expressed 57 pairwise preferences between the summaries generated by SummaRuNNer and \textsc{BanditSum}. The model with the lower score is better.}
\label{table:human_eval_summarunner}
\end{table}

\begin{table}[h!]
\small
\begin{center}
\begin{tabularx}{\columnwidth}{|l|X|X|X|}
\hline
Model & Overall & Coverage & Non-Redundancy \\ \hline
Refresh & 1.53 &\textbf{1.34}  & 1.55  \\ \hline
\B & \textbf{1.50} & 1.58 & \textbf{1.30} \\ \hline
\end{tabularx}
\end{center}
\caption{Average rank of manual evaluation with 4 participants who expressed 20 pairwise preferences between the summaries generated by Refresh and our system. The model with the lower score is better.}
\label{table:human_eval_refresh}
\end{table}

To compare with SummaRuNNer, we randomly sample 57 documents from the test set of DailyMail and ask 5 volunteers to evaluate the extracted summaries.
While comparing with Refresh, we use the 20 documents (10 CNN and 10 DailyMail) provided by \citet{DBLP:Narayan/2018} to 4 volunteers. Tables \ref{table:human_eval_summarunner} and \ref{table:human_eval_refresh} show the results of human evaluation in these two settings. \B is shown to be better than Refresh and SummaRuNNer in terms of overall quality and non-redundancy. These results indicate that the use of the true policy gradient, rather than the approximation used by Refresh, improves overall quality. It is interesting to observe that, even though \B does not have an explicit redundancy avoidance mechanism, it actually outperforms the other systems on non-redundancy.

\subsection{Learning Curve}
\begin{figure}[!h]
  \includegraphics[width=0.5\textwidth]{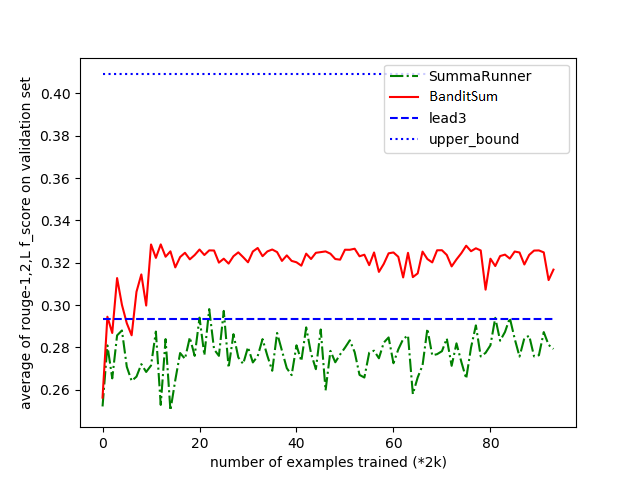}
  \caption[Caption for LOF]{Average of ROUGE-1,2,L F1 scores on the Daily Mail validation set within one epoch of training on the Daily Mail training set. The x-axis (multiply by 2,000) indicates the number of data example the algorithms have seen. The supervised labels in SummaRuNNer are used to estimate the \texttt{upper\_bound}.}
  \label{fig:train_efficiency}
\end{figure}

Reinforcement learning methods are known for sometimes being unstable during training. However, this seems to be less of a problem for \Bnospace, perhaps because it is formulated as a contextual bandit rather than a sequential labeling problem. We show this by comparing the validation curves generated by \B and the state-of-the-art maximum likelihood-based model -- SummaRuNNer \citep{ext5_summarunner} (Figure~\ref{fig:train_efficiency}).

From Figure~\ref{fig:train_efficiency}, we observe that \B converges significantly more quickly to good results than SummaRuNNer. Moreover, there is less variance in the performance of \Bnospace. One possible reason is that extractive summarization does not have well-defined supervised labels. There exists a mismatch between the provided labels and human-generated abstractive summaries. Hence, the gradient, computed from the maximum likelihood loss function, is not optimizing the evaluation metric of interest. Another important message is that both models are still far from the estimated upper bound\footnote{The supervised labels for the \texttt{upper\_bound} estimation are obtained using the heuristic described in \citet{ext5_summarunner}.}, which shows that there is still significant room for improvement.



\subsection{Run Time}
On CNN/Daily mail dataset, our model's time-per-epoch is about 25.5 hours on a TITAN Xp. We trained the model for 3 epochs, which took about 76 hours in total. For comparison, DQN took about 10 days to train on a GTX 1080 \citep{yao2018deep}. Refresh took about 12 hours on a single GPU to train \citep{DBLP:Narayan/2018}. Note that this figure does not take into account the significant time required by Refresh for pre-computing ROUGE scores.


\section{Discussion: Contextual Bandit Setting Vs. Sequential Full RL Labeling \label{sec:discussion}}
We conjecture that the contextual bandit (CB) setting is a more suitable framework for modeling extractive summarization than the sequential binary labeling setting, especially in the cases when good summary sentences appear later in the document. The intuition behind this is that models based on the sequential labeling setting are affected by the order of the decisions, which biases towards selecting sentences that appear earlier in the document. By contrast, our CB-based RL model has more flexibility and freedom to explore the search space, as it samples the sentences without replacement based on the affinity scores. Note that although we do not explicitly make the selection decisions in a sequential fashion, the sequential information about dependencies between sentences is implicitly embedded in the affinity scores, which are produced by bi-directional RNNs.  
 


\newcommand{\Dearly}{\mathit{D}_{\textit{early}}}
\newcommand{\Dlate}{\mathit{D}_{\textit{late}}}

\begin{figure}[!h]
  \includegraphics[width=0.5\textwidth]{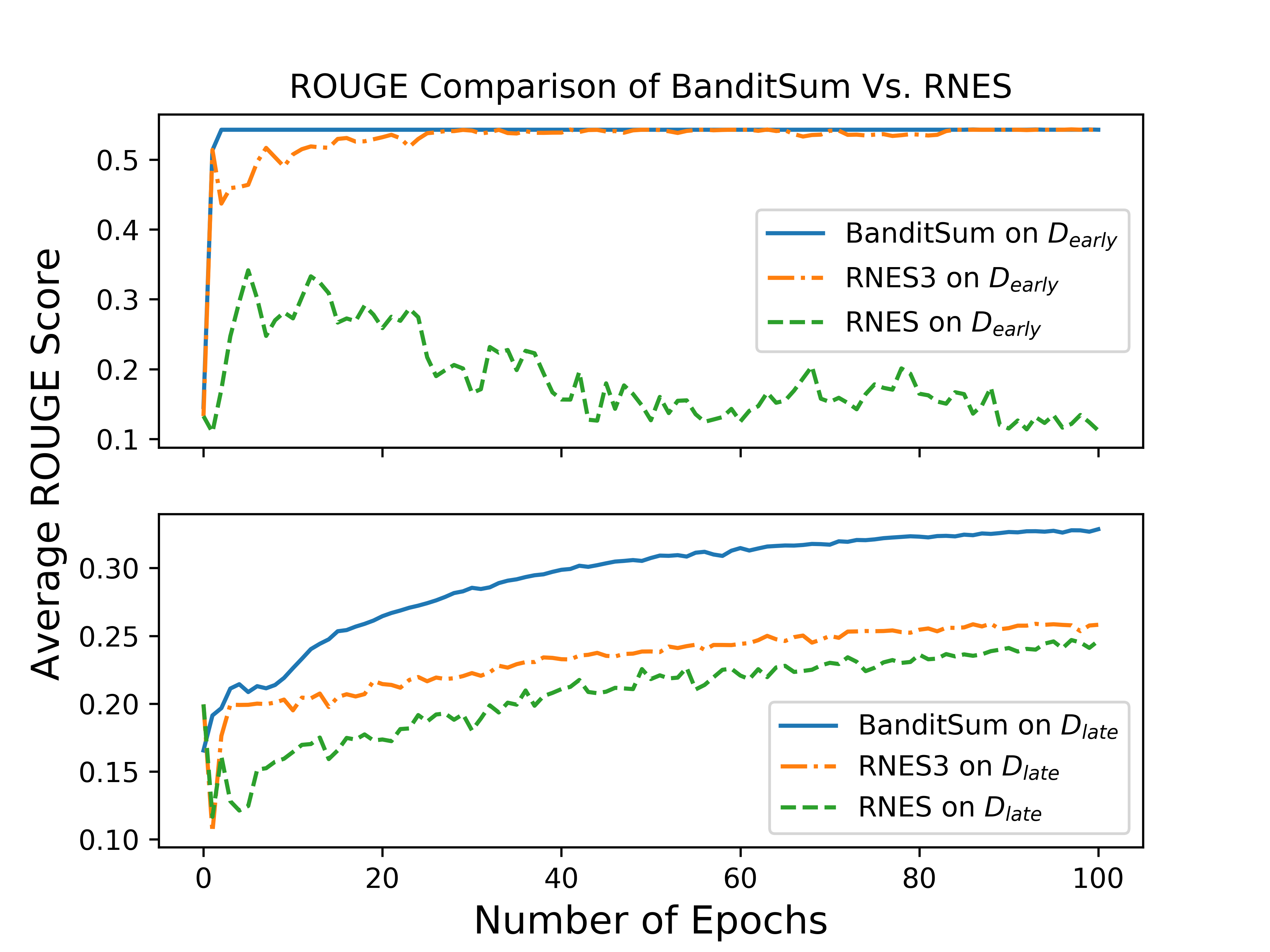}
  \caption{Model comparisons of the average value for ROUGE-1,2,L F1 scores ($\overline{f}$) on $\Dearly$ and $\Dlate$. For each model, the results were obtained by averaging $\overline{f}$  across ten trials with 100 epochs in each trail. $\Dearly$ and  $\Dlate$ consist of 50 articles each, such that the good summary sentences appear early and late in the article, respectively. We observe a significant advantage of \B compared to RNES and RNES3 (based on the sequential binary labeling setting) on $\mathit{D}_{late}$.}
  \label{fig:model_comparison}
\end{figure}

We provide empirical evidence for this conjecture by comparing \B to the sequential RL model proposed by~\citet{DBLP:conf/aaai/WuH18} (Figure \ref{fig:model_comparison}) on two subsets of the data: one with good summary sentences appearing early in the article, while the other contains articles where good summary sentences appear late. Specifically, we construct two evaluation datasets by selecting the first 50 documents ($\Dearly$, i.e., best summary occurs early) and the last 50 documents ($\Dlate$, i.e., best summary occurs late) from a sample of 1000 documents that is ordered by the average extractive label index \textit{$\overline{idx}$}. Given an article with $n$ sentences indexed from $1,\ldots, n$ and a greedy extractive labels set with three sentences $(i,j,k)$\footnote{For each document, a length-$3$ extractive summary with near-optimal ROUGE score is selected following the heuristic proposed by \citet{ext5_summarunner}.},  the average index for the extractive label is computed by \textit{$\overline{idx}$}$ = (i+j+k)/3n$.

Given these two subsets of the data, three different models (\Bnospace, RNES and RNES3) are trained and evaluated on each of the two datasets without extractive labels. Since the original sequential RL model (RNES) is unstable without supervised pre-training, we propose the RNES3 model that is limited to select no more then three sentences. Starting with random initializations without supervised pre-training, we train each model ten times for 100 epochs and plot the learning curve of the average ROUGE-F1 score computed based on the trained model in Figure \ref{fig:model_comparison}. We can clearly see that \B finds a better solution more quickly than RNES and RNES3 on both datasets. Moreover, it displays a significantly speed-up in the exploration and finds the best solution when good summary sentences appeared later in the document ($\Dlate$).

\section{Conclusion \label{sec:conclusion}}
In this work, we presented a contextual bandit learning framework, \B, for extractive summarization, based on neural networks and reinforcement learning algorithms. \B does not require sentence-level extractive labels and optimizes ROUGE scores between summaries generated by the model and abstractive reference summaries. Empirical results show that our method performs better than or comparable to state-of-the-art extractive summarization models which must be pre-trained on extractive labels, and converges using significantly fewer update steps than competing approaches. In future work, we will explore the direction of adding an extra coherence reward \citep{DBLP:conf/aaai/WuH18} to improve the quality of extracted summaries in terms of sentence discourse relation. 


\section*{Acknowledgements}
The research was supported in part by Natural Sciences and Engineering Research Council of Canada (NSERC). The authors would like to thank Compute Canada for providing the computational resources.


\bibliography{emnlp2018}
\bibliographystyle{acl_natbib_nourl}
\onecolumn
\section*{\textsc{BanditSum}: Extractive Summarization as a Contextual Bandit\\Supplemental Materials}
\appendix

\section{Samples}
\begin{table}[h!]
\begin{center}
\begin{tabularx}{\columnwidth}{|X|}
\hline
\textbf{Source Document} \\ 
\hline
\textcolor{blue}{(CNN)A top al Qaeda in the Arabian Peninsula leader -- who a few years ago was in a U.S. detention facility -- was among five killed in an airstrike in Yemen, the terror group said, showing the organization is vulnerable even as Yemen appears close to civil war.}
\textcolor{red}{Ibrahim al-Rubaish died Monday night in what AQAP's media wing, Al-Malahem Media, called a "crusader airstrike."} 
The Al-Malahem Media obituary characterized al-Rubaish as a religious scholar and combat commander.
\textcolor{yellow}{A Yemeni Defense Ministry official and two Yemeni national security officials not authorized to speak on record confirmed that al-Rubaish had been killed, but could not specify how he died.}
\textcolor{green}{Al-Rubaish was once held by the U.S. government at its detention facility in Guantanamo Bay, Cuba.} 
In fact, he was among a number of detainees who sued the administration of then-President George W. Bush to challenge the legality of their confinement in Gitmo.
He was eventually released as part of Saudi Arabia's program for rehabilitating jihadist terrorists, a program that U.S. Sen. Jeff Sessions, R-Alabama, characterized as "a failure." In December 2009, Sessions listed al-Rubaish among those on the virtual " 'Who's Who' of al Qaeda terrorists on the Arabian peninsula ... who have either graduated or escaped from the program en route to terrorist acts."
The United States has been active in Yemen, working closely with governments there to go after AQAP leaders like al-Rubaish. While it was not immediately clear how he died, drone strikes have killed many other members of the terrorist group.
Yemen, however, has been in disarray since Houthi rebels began asserting themselves last year. The Shiite minority group even managed to take over the capital of Sanaa and, in January, force out Yemeni President Abdu Rabu Mansour Hadi -- who had been a close U.S. ally in its anti-terror fight.
Hadi still claims he is Yemen's legitimate leader, and he is working with a Saudi-led military coalition to target Houthis and supporters of former President Ali Abdullah Saleh.
Meanwhile, Yemen has been awash in violence and chaos -- which in some ways has been good for groups such as AQAP. A prison break earlier this month freed 270 prisoners, including some senior AQAP figures, according to a senior Defense Ministry official, and the United States pulled the last of its special operations forces out of Yemen last month, which some say makes things easier for AQAP.\\ 
\hline
\textbf{Reference} \\
\hline
AQAP says a ``crusader airstrike'' killed Ibrahim al-Rubaish. Al-Rubaish was once detained by the United States in Guantanamo. \\
\hline
\textbf{Refresh}  \\ 
\hline
\textcolor{blue}{(CNN) A top al Qaeda in the Arabian Peninsula leader -- who a few years ago was in a U.S. detention facility -- was among five killed in an airstrike in Yemen, the terror group said, showing the organization is vulnerable even as Yemen appears close to civil war.} 
\textcolor{red}{Ibrahim al-Rubaish died Monday night in what AQAP's media wing, Al-Malahem Media, called a ``crusader airstrike.''} 
\textcolor{green}{Al-Rubaish was once held by the U.S. government at its detention facility in Guantanamo Bay, Cuba.} \\
\hline
\textbf{\B} \\ 
\hline
\textcolor{red}{Ibrahim al-Rubaish died Monday night in what AQAP 's media wing, Al-Malahem Media, called a ``crusader airstrike.''}
\textcolor{yellow}{A Yemeni Defense Ministry official and two Yemeni national security officials not authorized to speak on record confirmed that al-Rubaish had been killed, but could not specify how he died.} 
\textcolor{green}{Al-Rubaish was once held by the U.S. government at its detention facility in Guantanamo Bay, Cuba.} \\
\hline
\end{tabularx}
\end{center}
\caption{Example from the dataset showing summaries generated by \B and Refresh. Refresh selected the article’s first sentence (blue) which contains the redundant information ``Ibrahim al-Rubaish was in a U.S. detention facility''. In contrast, BanditSum selected the yellow sentence containing extra source information.}
\label{table:sample1}
\end{table}

\begin{table}[h]
\begin{center}
\begin{tabularx}{\columnwidth}{|X|}
\hline
\textbf{Source Document} \\ 
\hline
(CNN)You probably never knew her name, but you were familiar with her work.
\textcolor{blue}{Betty Whitehead Willis, the designer of the iconic "Welcome to Fabulous Las Vegas" sign, died over the weekend.} 
\textcolor{red}{She was 91.}
\textcolor{yellow}{Willis played a major role in creating some of the most memorable neon work in the city.}
\textcolor{green}{The Neon Museum also credits her with designing the signs for Moulin Rouge Hotel and Blue Angel Motel.}
\textcolor{orange}{Willis visited the Neon Museum in 2013 to celebrate her 90th birthday.}
Born about 50 miles outside of Las Vegas in Overton, she attended art school in Pasadena, California, before returning home.
She retired at age 77.
Willis never trademarked her most-famous work, calling it "my gift to the city."
Today it can be found on everything from T-shirts to refrigerator magnets.
\\
\hline
\textbf{Reference} \\
\hline
Willis never trademarked her most-famous work, calling it ``my gift to the city''. She created some of the city's most famous neon work. \\
\hline
\textbf{Refresh}  \\ 
\hline
\textcolor{blue}{Betty Whitehead Willis, the designer of the iconic ``Welcome to Fabulous Las Vegas'' sign, died over the weekend.}
\textcolor{yellow}{Willis played a major role in creating some of the most memorable neon work in the city.} 
\textcolor{orange}{Willis visited the Neon Museum in 2013 to celebrate her 90th birthday.} \\
\hline
\textbf{\B} \\ 
\hline
\textcolor{blue}{Betty Whitehead Willis , the designer of the iconic ``Welcome to Fabulous Las Vegas'' sign , died over the weekend.} 
\textcolor{red}{She was 91.} 
\textcolor{green}{The Neon Museum also credits her with designing the signs for Moulin Rouge Hotel and Blue Angel Motel.} \\
\hline
\end{tabularx}
\end{center}
\caption{Example from the dataset showing summaries generated by \B and Refresh. \B tends to select more precise information.}
\label{table:sample1}
\end{table}

\end{document}